\documentclass[letterpaper, hidelinks, 10 pt, journal, twoside]{IEEEtran}

\IEEEoverridecommandlockouts

\usepackage[amssymb]{SIunits}
\usepackage{xcolor}
\usepackage{cite}
\usepackage{epsfig}
\usepackage{graphicx}
\graphicspath{ {images/} }
\usepackage{amsfonts}
\usepackage{amsmath}
\usepackage{svg}
\usepackage{amsmath}
\usepackage{makecell}
\usepackage{subfigure}
\usepackage{mathtools}
\usepackage{placeins}
\usepackage{color}
\usepackage{soul}
\usepackage{booktabs}
\usepackage{color,soul}
\usepackage{svg}
\usepackage{float}
\usepackage{booktabs}
\usepackage{graphics}
\usepackage{algorithm}
\usepackage[nodisplayskipstretch]{setspace}
\usepackage[normalem]{ulem}
\useunder{\uline}{\ul}{}
\usepackage{booktabs}
\usepackage[switch]{lineno}
\usepackage{url}
\usepackage[inline]{enumitem}

\begin{document}

\title{\LARGE \bf
This is the Way:\\Differential Bayesian Filtering for Agile Trajectory Synthesis
}
\author{Trent Weiss and Madhur Behl \\
Department of Computer Science \\
University of Virginia, Charlottesville, USA \\
\{ttw2xk \& madhur.behl\}@virginia.edu
}
\markboth{IEEE Robotics and Automation Letters. Preprint Version. Accepted July, 2022}{}

\maketitle

\begin{abstract}
% Autonomous vehicles with the ability to execute agile racing maneuvers have recently received a lot of attention from the robotics and machine learning research communities. 
One of the main challenges in autonomous racing is to design algorithms for motion planning at high speed, and across complex racing courses. 
End-to-end trajectory synthesis has been previously proposed where the trajectory for the ego vehicle is computed based on camera images from the racecar.  
This is done in a supervised learning setting using behavioral cloning techniques. In this paper, we address the limitations of behavioral cloning methods for trajectory synthesis by introducing Differential Bayesian Filtering (DBF), which uses probabilistic B\'ezier curves as a basis for inferring optimal autonomous racing trajectories based on Bayesian inference. We introduce a trajectory sampling mechanism and combine it with a filtering process which is able to push the car to its physical driving limits. The performance of DBF is evaluated on the DeepRacing Formula One simulation environment and compared with several other trajectory synthesis approaches as well as human driving performance. DBF achieves the fastest lap time, and the fastest speed, by pushing the racecar closer to its limits of control while always remaining inside track bounds.
\vspace{-2mm}
\end{abstract}

\IEEEpeerreviewmaketitle

\section*{SUPPLEMENTARY VIDEO}
This paper is accompanied by a narrated video of the
performance: \url{https://youtu.be/5SezU0ICaug}

\setlength{\textfloatsep}{1.0pt}
\section{Introduction}
\label{sec:intro}
In motorsport racing, there is a saying that \emph{If everything seems under control, then you are not going fast enough}. 
Expert racing drivers have split second reaction times and routinely drive at the limits of control, traction, and agility of the racecar - under high-speed and close proximity situations. 
Autonomous racing presents unique opportunities and challenges in designing algorithms that can operate firmly on the limits of perception, planning, and control.

While autonomous vehicle research and development is focused on handling routine driving situations, achieving the safety benefits of autonomous vehicles also requires a focus on driving at the limits of the control of the vehicle.  In recent years autonomous racing competitions, such as F1/10 autonomous racing~\cite{o2019f1,babu2020f1tenth} and the Indy Autonomous Challenge~\cite{indy} are becoming proving grounds for testing motion planning, and control algorithms at high speeds.
%Roborace~\cite{roborace}, 
With the autonomous racing application in mind, this paper focuses on the problem of trajectory synthesis or motion planning for an autonomous racecar. In our previous work~\cite{deepracing-date}, we have demonstrated end-to-end autonomous racing in the widely popular Formula One (F1) game used by real F1 drivers. 
Deep learning based approaches for trajectory synthesis for autonomous vehicles have either been one where a trajectory is predicted via a prediction of future waypoints for the ego vehicle to follow, or in some cases a parameterized curve is predicted instead and waypoints are sampled from the curve~\cite{chen2015deepdriving,codevilla2018end,muller2018driving}. In either case, these techniques still rely on supervised learning from expert behavior for predicting the trajectory of the ego vehicle. Once the predicted trajectory is computed a low-level controller such as pure-pursuit or a model predictive control can follow that trajectory.%{\color{red}}

\begin{figure}
    \centering
    \includegraphics[width=\columnwidth]{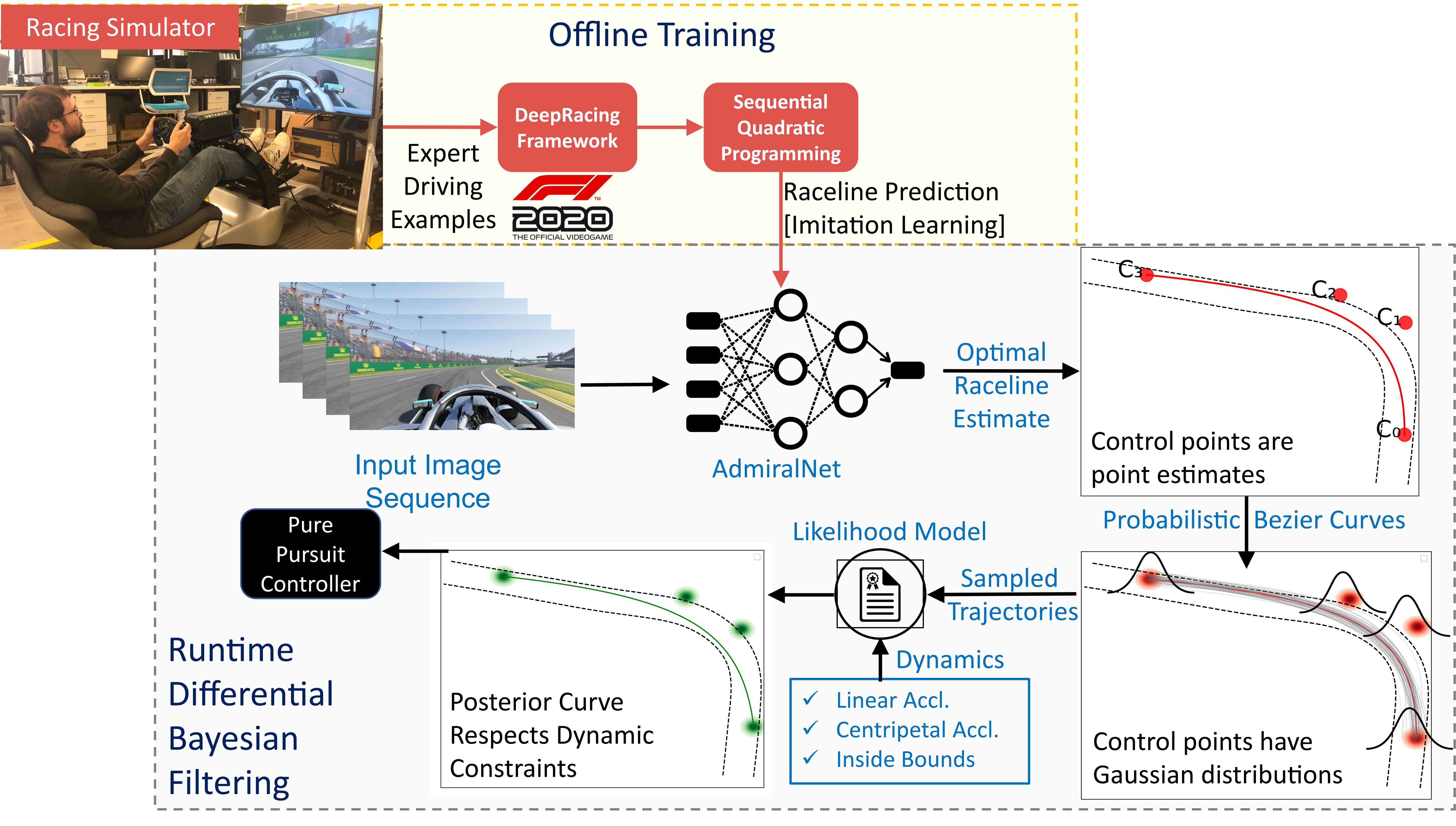}
    \caption{The key idea in Differential Bayesian Filtering is that we fit Gaussian distributions over a B\'ezier curve's control points to generate a distribution of trajectories that respect the ego vehicle's dynamic limits.}
    \label{fig:overview}
    % \vspace{-6mm}
\end{figure}

However, behavior cloning based trajectory 
synthesis is brittle~\cite{codevilla2019exploring} because at best it can generate trajectories which are averaged from the input/training data.
%by observing expert racing drivers. 
These methods do not generalize to computing trajectories which respect vehicle dynamics, or track bounds at all times. Nor can they leverage the knowledge of vehicle dynamics.
%They also suffer a domain shift between the off-line training experience and the on-line behavior~\cite{ross2011reduction}.

In this paper, we address major limitations of supervised learning methods for trajectory synthesis for autonomous racing by introducing a new method called Differential Bayesian Filtering (Fig.~\ref{fig:overview}). This is a significant improvement over previous methods for trajectory synthesis, and can be summed up with 3 major contributions:

\begin{enumerate}%[labelsep=*,leftmargin=0.85pc]
    \item The ability to learn distributions over the space of desired trajectories.
    \item A Monte-Carlo sampling method for inferring closer-to-optimal trajectories from those distributions
    \item A filtering method to combine deep learning trajectory synthesis with knowledge of the vehicle's feasibility limits.
\end{enumerate}
% {\color{red}}
We show empirically in Section \ref{sec:results} that our novel framework, built on top of Probabilistic B\`ezier Curves and Monto-Carlo based Bayesian inference, generates trajectories which are outside of the training data distribution and results in significantly better performance than the best human driving example in the training data.% {\color{red} }
We select autonomous racing as the application domain for our work, but it can generalize to motion planing for any autonomous vehicle. %We now describe related work in this domain.
% Using our method, we can learn distributions over the input trajectory space and then sample optimal trajectories from those distributions. 
% This results in a more stable and faster autonomous racing behavior, which we demonstrate using our DeepRacing simulation testbed. 
% 

\section{Related Work}
\label{sec:related_work}
There is existing work on trajectory synthesis for autonomous driving \cite{rttp,dtp,Astar}. However, these works only consider a ``$0^{th}$" order view of the problem i.e. planning a trajectory based only on the ego vehicle's position. They do not consider any differential or higher order constraints which are desirable for speed and acceleration behavior.

% Approaches based on trajectory optimization are also quite popular~\cite{lattice_searching,convex_optim,optim_1_43}.  However, these techniques are highly sensitive to the ``initial guess" used.%  Our work addresses this limitation by using the output of neural network as as prior distribution for Bayesian inference.

In~\cite{plop}, authors present a probabilistic approach for trajectory synthesis which predicts a $4^{th}$ order spline.  
However,~\cite{plop} does not consider distributions over the space of polynomials, as we do, but uses Gaussians over fixed points along a predicted polynomial as a convenient means of constructed their network's loss function. 
Additionally,~\cite{plop} only uses behavioral cloning which is not optimal and does not align with the goal of pushing the vehicle to its limits. 
%Finally,~\cite{plop} not consider any driving performance metrics in their evaluation, such as lap time/average speed/etc., and only reports difference from the human expert in their experiments. 
%and is not aligned with our stated goal of producing an \emph{agile} racing stack.
%In~\cite{plop}, authors use a $4^{th}$ order spline parameterization that is only expressable at fixed points on the curve. Also, the PLOP model only considers predicted position vs ground-truth in it's loss function. Our approach views the problem as one of Bayesian inference on the parameters of the curve itself, rather than points on that curve. Because a B\`ezier curve's derivatives are just linear combinations of it's parameters (control points), this enables inference on differential properties of the optimal curve.
% Additionally, the authors in~\cite{plop} do not control the ego vehicle with the trajectories they predict.  Consequently, their ``closed-loop" evaluation is incomplete, as they are not actually controlling the vehicle, but are only evaluation how close their predicted trajectory is to a human example while an oracle agent controls the vehicle.  We show that our method works in truly closed-loop fashion, i.e. a fully function autonomous racing stack.

Researchers have looked at the problem of determining an optimal raceline~\cite{minimum_curvature_trajectories, raceline_bayesian_optim}. 
In these works, the raceline is determined by an offline optimization that can incorporate knowledge of vehicle dynamics. Our method, on the other hand, is \emph{online} and infers optimal behavior from desirable characteristics of the vehicle's motion at runtime.  
We aren't proposing a new method to find the optimal raceline for a track, rather we are proposing a runtime method, that generates fast trajectories for a racecar - which converges to the optimal raceline when the racecar runs by itself but can also be extended to multi-agent racing situations. %{\color{red} }
% This makes our method flexible and extensible to dynamic environments where these desirable characteristics can change with time, such as multi-agent scenarios or situations where the drivable area has changed, e.g. a disabled vehicle on-track that must be avoided.}
%These are motivated by a similar problem as ours, the need to produce curves with optimal differential properties.  We build on these works with a Bayesian interpretation of the optimal racing line under a convenient parameterization.

%There is a large body of work in end-to-end autonomous driving. {\color{red}}
In~\cite{event-frames,memory-cells,discrete-action-model}, the authors use fully end-to-end style architectures.  I.e. images from the driver's point of view are mapped directly to control outputs for the car (steering and throttle).  However, this approach is very brittle and usually does not generalize well to out-of-distribution images.  Other work~\cite{LSTM-highway,deep-path-planning} remedies this limitation by using neural networks to predict a series of waypoints for the autonomous vehicle to follow.  Waymo's ChauffeurNet~\cite{ChauffeurNet} learns to follow specified waypoints and specified speeds.  The survey in~\cite{le_mero_survey} also lists several end-to-end architectures for autonomous driving. Most use some form of Convolutional Neural Networks, sometimes integrated with LSTM cells or another form of recurrent network. For example, in~\cite{cai_vision_based_learning}, the authors utilize a CNN-LSTM architecture for predicting a fixed number of waypoints and velocities for the car to follow, similiar to ~\cite{ChauffeurNet}.
%However, this strategy limits opportunity for trajectory optimization as the derivative information for the predicted path is lost by the waypoint encoding. 
However, predicting sufficient waypoints leads to the curse of dimensionality as the neural network needs to predict proportionally more parameters to specify more waypoints. In our method, we only need to predict the control points of a parameterized Bezier curve, which fixes the dimensionality of our problem and allows implicit encoding of the predicted trajectories derivatives.
%%
%Reinforcement Learning is also a common approach to this problem.
~\cite{gran_turismo_rl} is an example of work which uses Deep Reinforcement Learning (DRL) to autonomously race in the Gran Turismo game. This is close to our work in terms of the problem setting but very different in terms of methodology.
DRL techniques require numerous exploration runs of the action space, including experiencing many crashes while learning to maximize racing related rewards.  
%This presents a significant challenge for deploying in the real world: repeated crashing during the training process would be prohibitively expensive on a real-world racecar.  
Our method requires only a few hours worth of training data and is able to push the car to its limits.  
Additionally, our Formula One DeepRacing framework does not require any special access to the internal state of the game engine - it is reproducible by anyone who owns the game unlike the Grand Turismo work which required changing game state to enable DRL.

We build upon our previous work in ~\cite{deepracing_bezier, neurips_deepracing_bezier} that uses a neural network to predict a B\`ezier curve as a canonical representation of a desired trajectory for the autonomous vehicle.  This model was shown to outperform waypoint prediction, and the parameterized representation is a convenient form for applying Bayesian inference techniques to estimate a more-optimal trajectory.  We next present a brief overview of probabilistic B\'ezier curves, an extension of B\`ezier curves to a Gaussian probability model, which will form the basis of our Bayesian method for agile trajectory synthesis. 

\section{Probabilistic B\`ezier Curves}
\label{sec:pbc}
A B\'ezier curve is a parametric curve used in computer graphics and related fields. 
The curve, a linear combination of Bernstein polynomials, is named after Pierre B\'ezier, who developed them to model Renault racecars.
%~\cite{lorentz2013bernstein}
A B\'ezier curve is formed from a combination of Bernstein polynomials (Eq~\ref{eqn:berstein_polynomial}) that maps a scalar parameter $s \in [0,1]$ to a point in a euclidean space of dimension $d$, $\mathbf{R}^d$.  More specifically, a B\'ezier curve is a polynomial combination of a set of ``control points".
%These control points define a B\'ezier curve as follows. %~\cite{mathematics-for-cg}
A B\'ezier curve of degree k isdefined by k+1 control points:
$\{ \mathbf{C}_0,  \mathbf{C}_1, \mathbf{C}_2,  ..., \mathbf{C}_k \in \mathbf{R}^d \} $.  The corresponding B\'ezier curve, $\mathbf{B} : [0,1] \rightarrow \mathbf{R}^d$, as a function of a unitless scalar, $s$, is:
% {\color{red}}
\begin{equation}
    \small
\label{eqn:berstein_polynomial}
    b_{i,k}(s) = \binom{k}{i}{(1-s)}^{k-i}s^i
\end{equation}
    % $$b_{i,k}(s) = \binom{k}{i}{(1-s)}^{k-i}s^i$$
\vspace{-2mm}
\begin{equation}
    \small
\label{eqn:bezier_eval}
    \mathbf{B}(s) = \sum_{i=0}^k b_{i,k}(s) \mathbf{C}_i
\end{equation}
Note that a B\'ezier curve always starts at $\mathbf{C}_0$ ($s=0$) and ends at $\mathbf{C}_k$ ($s=1$). A B\'ezier curve's derivative w.r.t $s$ is:
\begin{equation}
\label{eqn:bezier_space_derivative}
    \small
    \frac{d\mathbf{B}}{ds} = k\sum_{i=0}^{k-1} b_{i,k-1}(s) (\mathbf{C}_{i+1} - \mathbf{C}_i)
\end{equation}

Under this formulation, the scalar parameter $s$ is just unitless parameter on $[0,1]$.  To express the curve as a function of \emph{time}, we specify a total amount of time, $\Delta t$, for the curve to go from $\mathbf{C}_0$ to $\mathbf{C}_k$. I.e. time, $t$, can be expressed as $t=s\Delta t$. Equivalently, $s=\frac{t}{\Delta t}$.  Under this formulation, the velocity vector, w.r.t time, of a B\'ezier curve is simply:
\begin{equation}
\label{eqn:ds_dt}
    \small
    \frac{ds}{dt}=\frac{1}{\Delta t}
\end{equation}
\begin{equation}
\label{eqn:bezier_time_derivative}
    \frac{d\mathbf{B}}{dt} = \frac{ds}{dt}\frac{d\mathbf{B}}{ds} = \frac{1}{\Delta t}\frac{d\mathbf{B}}{ds}
\end{equation}

$\Delta t$ determines how fast the curve is as a function of time, smaller $\Delta t$ implies a faster curve and a larger $\Delta t$ implies a slower curve.  How this chosen constant determines the velocity of the curve will be important for our Bayesian formulation of path planning.

Hug et al.~\cite{probabilistic_bcurves} introduced the concept of probabilistic B\'ezier curves.
%as a Gaussian Distribution over the space of B\`ezier curves.  
A probabilistic curve is defined not by a fixed set of control points, but by a set of mutually-independent Gaussian distributions over each control point, where each Gaussian vector is defined by a mean, $\mathbf{C}_i \in \mathbf{R}^d$, and a covariance matrix, $\mathbf{\Sigma}_i \in \mathbf{R}^{d \times d}$.  I.e. a probabilistic B\'ezier curve, $\mathcal{B}(s)$, is defined by:
%\ \mathcal{C}_1\sim\mathcal{N}(\mathbf{C}_1,\mathbf{\Sigma}_1)
\begin{equation}
    \{ \mathcal{C}_0\sim\mathcal{N}(\mathbf{C}_0,\mathbf{\Sigma}_0),\ ...\ ,\ \mathcal{C}_k\sim\mathcal{N}(\mathbf{C}_k,\mathbf{\Sigma}_k) \}
    \label{eqn:pbc_set}
\end{equation}
\begin{equation}
    \mathcal{B}(s)\sim\sum_{i=0}^k b_{i,k}(s) \mathcal{C}_i
    \label{eqn:pbc_equation}
\end{equation}
Where each $\mathcal{C}_i \sim \mathcal{N}(\mathbf{C}_i,\mathbf{\Sigma}_i)$ is independent of the others.
%{\color{red}}

% In~\cite{probabilistic_bcurves}, the authors show that, given a probabilistic B\`ezier curve,  $\mathcal{P}$, the resulting distribution of any point on that curve is also Gaussian with the following distribution:

% $$ \mathcal{B}(s) \sim \mathcal{N}(\sum_{i=0}^k b_{i,k}(s)\mathbf{C}_i, \sum_{i=0}^k {b_{i,k}(s)}^2\mathbf{\Sigma}_i)$$

% The proof is briefly recounted here.  Each control point $\mathcal{C}_i$ is mutually independent of the others.  Therefore, $\mathcal{B}(s)$ is a linear combination of Gaussian random vectors.  Recall the following property of linear combinations of Gaussian random vectors. If $\{\mathcal{X}_0 \sim \mathcal{N}(\mathbf{u}_0, \mathbf{\Sigma}_0), \mathcal{X}_1 \sim \mathcal{N}(\mathbf{u}_1, \mathbf{\Sigma}_1), ... , \mathcal{X}_n \sim \mathcal{N}(\mathbf{u}_n, \mathbf{\Sigma}_n)\}$ are mutually independent Gaussian random vectors, then a linear combination of those random vectors is as follows.
% $$ \mathcal{Y} = \sum_{i=0}^n a_i \mathcal{X}_i$$
% $$ \mathcal{Y} \sim \mathcal{N}(\sum_{i=0}^n a_i\mathbf{u}_i, \sum_{i=0}^n {a_i}^2\mathbf{\Sigma}_i)$$

% Because a probabilistic B\`ezier curve is a linear combination of it's mutually independent control points, with the linear coefficients being each $b_{k,n}(s)$, it readily follows that 
% $$ \mathcal{B}(s) \sim \mathcal{N}(\sum_{i=0}^k b_{i,k}(s)\mathbf{C}_i, \sum_{i=0}^k {b_{i,k}(s)}^2\mathbf{\Sigma}_i)$$.

\section{Problem Formulation}
\label{sec:problem}
\begin{figure}[]
    \centering
    \includegraphics[width=0.775\columnwidth]{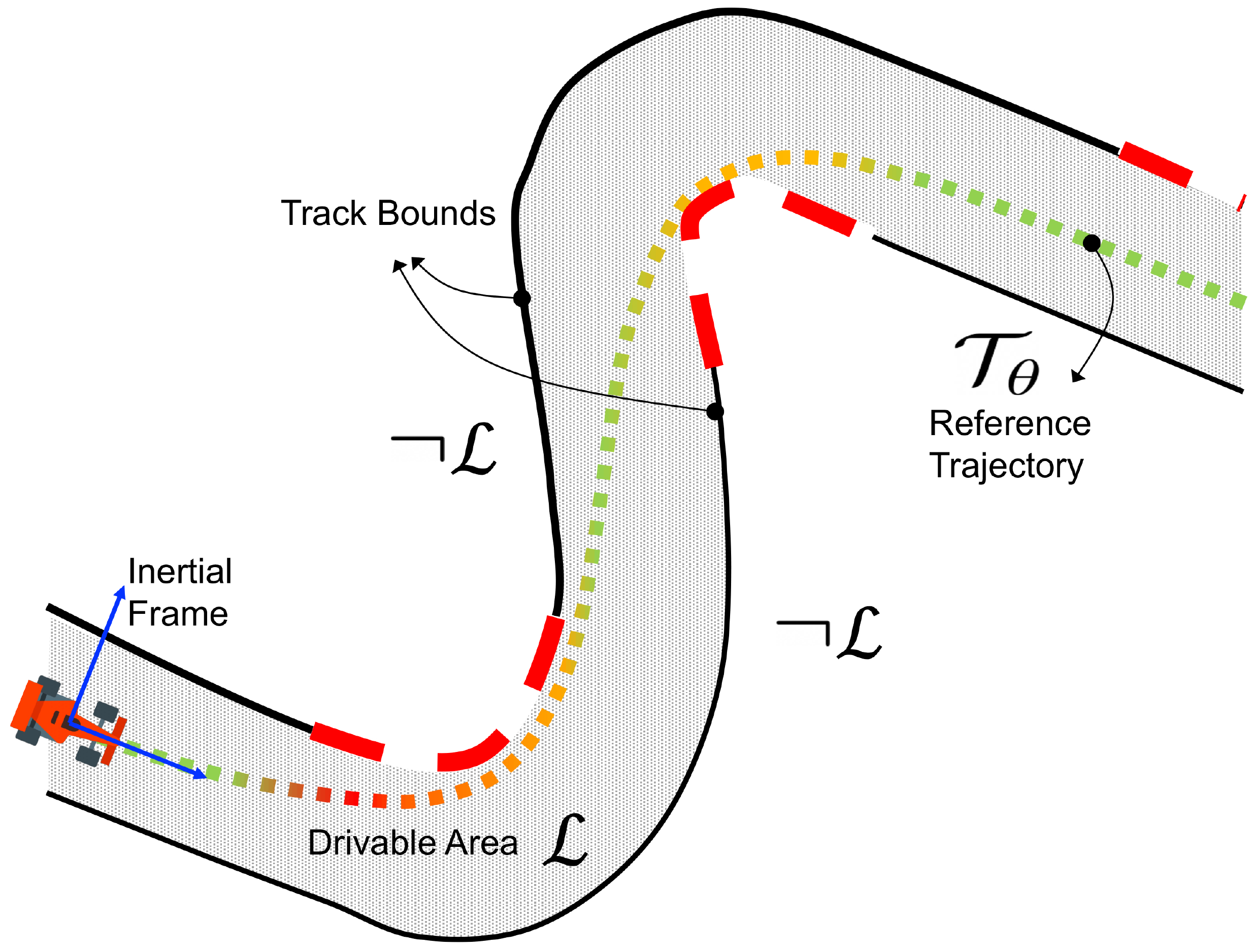}
    \caption{High-level view of autonomous racing.  The driveable and undrivable areas are $\mathcal{L}$ and $\neg\mathcal{L}$, respectively. We present a Bayesian approach to determining an optimal trajectory, $\mathcal{T}_{\theta}$, for the ego vehicle to follow. }
    \label{fig:problem}
    % \vspace{-6mm}
\end{figure}

This work is focused on a Bayesian interpretation of trajectory synthesis for autonomous racing. Given some sensor inputs, an autonomous racing agent needs to generate a trajectory to follow that represents desirable racing behavior, where ``desirable" means a trajectory that is as fast as possible without violating any kinematic/dynamic constraints of the vehicle or going outside the boundaries of the track.  In this work, a \textit{trajectory}, which we denote as $\mathcal{T}$, is a smooth curve in the ego vehicle's task space: $\mathbf{R}^2$. $\mathcal{T}$ is a set of points that can be expressed as a $C^{\infty}$ function of time, such that the terms velocity and acceleration have their intuitive meaning: the first and second time derivatives of the curve, respectively.  For this work, we focus on trajectories that can be fully described by a set of parameters: $\theta$. We denote such a parameterized trajectory as $\mathcal{T}_{\theta}$. For this work, we take $\theta$ to be the control points of a B\`ezier Curve. Figure~\ref{fig:problem} gives a high-level overview of this problem setting. An autonomous racing agent needs to select a parameterized trajectory that is as fast as possible without leaving the drivable area or exceeding the car's physical limits.  Once a trajectory is selected, a classical path-following algorithm can be used to decide on steering and throttle commands for the ego vehicle.  In this work, we select a Pure Pursuit controller \cite{pure_pursuit}, but other path-following algorithms (such as a Model Predictive Controller) could be used - the choice is agnostic to our framework. 
%{\color{red} }{\color{red}}{\color{red}}
The optimal racing line (ORL), $\mathcal{T^*}$, is the trajectory with maximal average velocity (over time, $t$) that is both achievable under the car's dynamic limits and is wholly contained within the track boundaries.  Let $\Lambda(\mathcal{T})$ be a boolean function that is true $\mathrm{iff}$ $\mathcal{T}$ is physically achievable. We denote the region within the track bounds with the symbol $\mathcal{L}$.  If $\dot{\mathcal{T}}$ and $\vert\dot{\mathcal{T}}\vert$ represent the velocity and speed, respectively, of $\mathcal{T}$, the optimal racing line is:
\begin{equation}
    \small    
    \mathcal{T^*} = \text{argmax}(\mathcal{T})\text{ } \mathbf{E}_{t}[\vert\dot{\mathcal{T}}\vert] : (\mathcal{T} \subset \mathcal{L})\text{ }\land \Lambda(\mathcal{T})
    \label{eqn:optimal_raceline_definition}
\end{equation}%Where $\dot{\mathcal{T}}$ represents the velocity of the curve $\mathcal{T}$ and $\vert\dot{\mathcal{T}}\vert$ is its speed.
% \vspace{-2mm}{\color{red}}
\begin{figure}%[!h]
    \centering
    \includegraphics[width=\columnwidth]{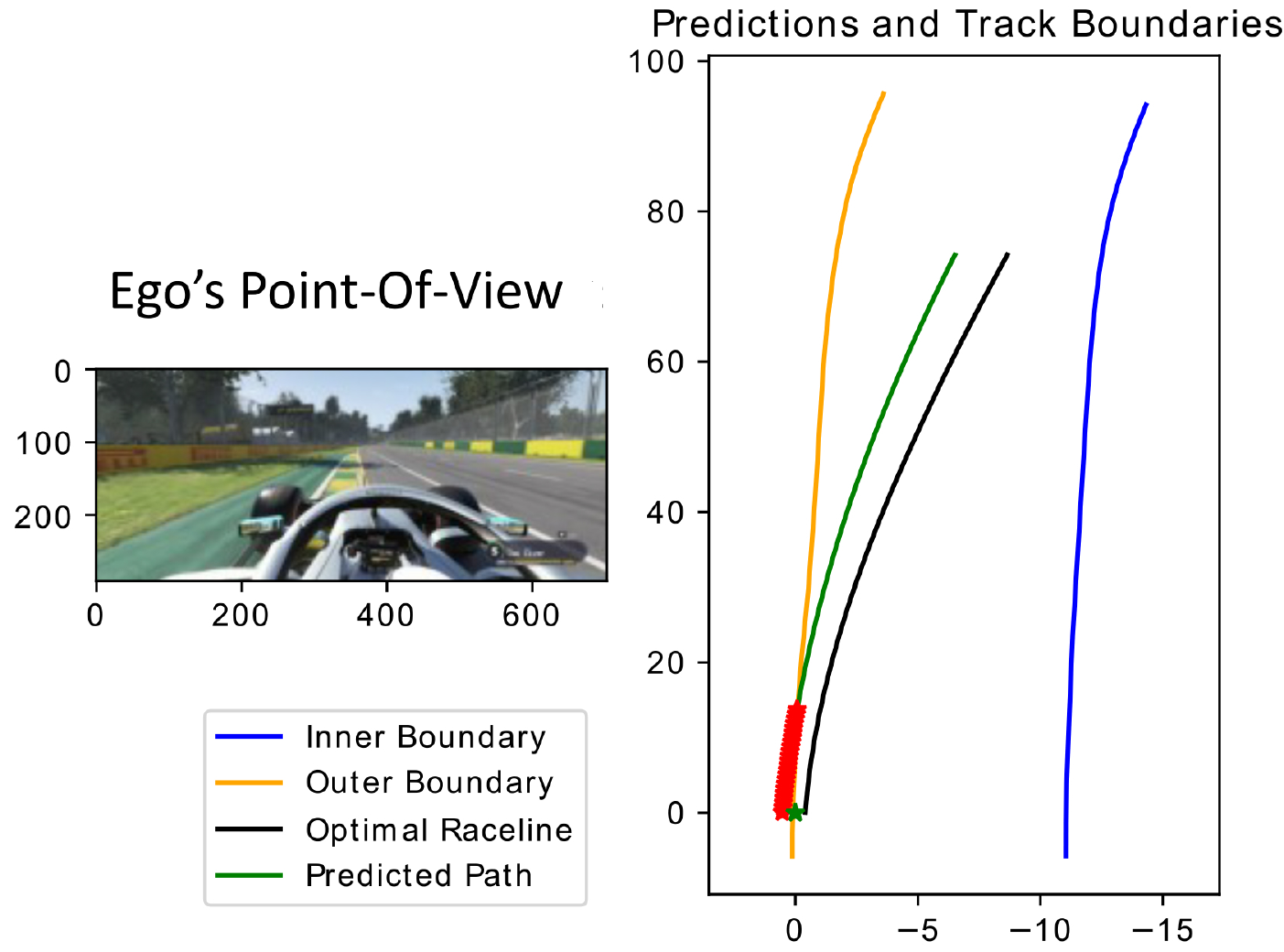}
    \caption{A neural network is trained the predict $\mathcal{T^*}$ (in black) 2.25$[\second]$ into the future based on images from the driver's point-of-view. However, it's predictions (in green) can be incorrect. We show that our Bayesian framework can mitigate the effect of incorrect network predictions and produce more optimal behavior.}
    \label{fig:pointestimates}
    % \vspace{-2mm}
\end{figure}% If an autonomous agent had ground-truth knowledge of this optimal racing line, an optimal control strategy would be to simply have the ego agent follow $\mathcal{T^*}$. There is existing work on computing $\mathcal{T^*}$ in a global (and fixed) coordinate system given the track boundaries in that coordinate system~\cite{optim_based_racing,raceline_bayesian_optim,minimum_curvature_trajectories}.  However, this task is computationally intensive and expressing this globally defined path in the ego vehicle's local coordinate system would require an accurate estimate of the ego vehicle's pose in that coordinate system.  This state estimation task can be difficult/computationally intensive for high-speed systems like a race car. An alternative approach is to generate a trajectory in the ego's local coordinate system at runtime which is close to the global optimal race line.The problem of autonomous racing can be thought of as inferring optimal behavior given imperfect knowledge \hl{how?- explain.} of the optimal racing line.  
We present a method of Bayesian inference, called Differential Bayesian Filtering, to estimate $\mathcal{T^*}$, which can then be passed to a path-following algorithm.  This technique is an extension of our previous work in~\cite{deepracing_bezier,neurips_deepracing_bezier} that uses a sequence of images as the input to a neural network, called AdmiralNet, that was trained to predict the control points of a B\`ezier curve approximation to $\mathcal{T^*}$.  For details on AdmiralNet's architecture, we refer the reader to~\cite{deepracing_bezier}.  However, our work in~\cite{deepracing_bezier} only provides a deterministic B\`ezier curve that might be incorrect for images that the neural network has never seen before (Fig.~\ref{fig:pointestimates}). To address this limitation, we use Bayesian inference for improving the point estimate by taking the output of that neural network as the mean of a probabilistic B\`ezier curve to serve as a Gaussian prior distribution for Bayesian inference.  
%This network is trained to predict a 2.25 second long piece of the global raceline. 
%We now describe how that optimal racing line is determined.  %  I.e. we take the control points of a B\`ezier curve as our parameterization, $\theta$, and $\mathcal{T}_\theta$ as the curve defined by those control points. We now describe our method of Bayesian Inference, called Differential Bayesian Filtering.

\section{Raceline Estimation}
\label{sec:raceline_optim}
We utilize Sequential Quadratic Programming (SQP) to specify the optimal racing line that the neural network is trained to predict. To start, we select the minimum curvature path that completes the entirety of the racetrack and stays within the track boundaries, a set of points, $P$: %Mathematically, this is a set of N points:{\color{red} }
\begin{equation}
    % \begin{align*}
    P = \overset{argmin}{[p_0, p_1, ..., p_{N-1}]}\text{ }\sum_{i=0}^{N-1} \kappa_i
    % \end{align*}
    \label{eqn:min_curvature_path}
\end{equation} % In this work, we study closed racetracks that start and end at the same point ($p_0=p_{N-1}$), but this is not required in general.
Where $[p_0, p_1, ..., p_{N-1}] \in \mathbb{R}^2$ traverses the entirety of the racetrack and $\kappa_i$ is the curvature of the path at $p_i$. However, this is just a set of points in $\mathbb{R}^2$.  To specify the optimal raceline as a trajectory (a function of time), we utilize SQP optimization to assign a speed at each point on the minimum curvature path, from which a time at each point readily follows.  We uniformly sample points at 1.5m intervals from the minimum curvature path , such that $v_i$ is the desired speed along the path at the point $p_i$.  We then impose two inequality constraints on the square of the car's speed ($v^2$) at each point:

First, we limit the centripetal (lateral) acceleration, $a_c$, of the car to 26.5 $[\meter\per\second\squared]$ ($\sim$2.7G). Real F1 cars experience $\sim3-4$G of centripetal acceleration, so we select this limit as a conservative estimate.
\begin{equation}
    \frac{{v^2}_i}{R_i}  \leq 26.5 [\meter\per\second\squared]\text{ } \forall i
    \label{eqn:centripetal_constraint}
\end{equation}%{\color{red} }
Where $R_i = \frac{1}{\kappa_i}$ is the radius of curvature of the path at $p_i$.   
 %{\color{red} } 
Second, we limit the longitudinal acceleration, $a_l$, of the car by assuming the car's longitudinal acceleration is constant between each $p_i$.  Let $\Delta r$ represent the path-length between any two adjacent points on the path and $a_l$ represent longitudinal (forward) acceleration. This constraint is then:
\begin{equation}
    \small    
    \setstretch{-1.5}
    % \vspace{-2mm}
    {v_i}^2 = {v_{i-1}}^2 + 2{a_{l_{i-1}}}\Delta r
    \label{eqn:delta_vsquare_accel}
\end{equation}
% \vspace{-1mm}
  Because we discretize the path at intervals of $\Delta r=1.5[\meter]$:
\begin{equation}
    \small    
    a_{min}(v_{i-1}) \leq  \frac{{v_i}^2 - {v_{i-1}}^2}{3} \leq  a_{max}(v_{i-1})
    \label{eqn:accel_constraint}
\end{equation}
% \vspace{-3mm}
% \begin{equation}
%     \frac{{v_i}^2 - {v_{i-1}}^2}{3} \geq  a_{min}(v_{i-1})
%     \label{eqn:braking_constraint}
% \end{equation}}
$a_{max}$ is a function that maps the car's speed to the maximum forward acceleration the car can achieve at that speed, which decreases as the car  goes faster due to drag. $a_{min}$ is a function that maps the car's speed to the fastest braking rate the car can achieve at that speed (a negative sign here indicates braking).  This will also decrease (braking is negative acceleration) as the car goes faster, because drag helps the car slow down more at higher speeds.  The closed form of $a_{max}$ and $a_{min}$ is very complex, as they both involve the complex aerodynamics of an F1 car body.  For this work, we determine $a_{min}$ and $a_{max}$ with step-response test on the car in the DeepRacing testbed and measured the maximum achievable acceleration and braking as a function of speed.  We use linear interpolation on these data as $a_{max}$ and $a_{min}$.
%However, this approach is extensible to physics-based limits on acceleration and braking derived from an aerodynamic model of the chassis and a model of the vehicle's power-train.}{\color{red}{\color{red}}{\color{red}}{\color{red}}{\color{red} }

Setting optimal velocities at each point then becomes a vector-space optimization problem.  Given a vector $\mathbf{v} = [{v^2}_0, {v^2}_1, ..., {v^2}_{N-1}]$, we determine velocities on the minimum-curvature path to be the solution to:
\begin{equation}
 \setstretch{0.5}
  \small
% \begin{step}
% \vspace{-3mm}
max \sum_{i=0}^{N-1}\mathbf{v}_i\text{  }s.t.\text{   }
\text{ }\mathbf{Av} \leq \boldsymbol{\gamma}_1 \text{,  }
 \mathbf{Bv} \leq \boldsymbol{\gamma}_2 \text{,  }
 \mathbf{Bv} \geq \boldsymbol{\gamma}_3
% \end{step}
\label{eqn:sqp_formulation}
\end{equation}
% \vspace{-2mm}
% Where 
% \begin{align*}
\begin{equation}
  \small
    % \mathbf{A}=
    % \begin{bmatrix}
    %     \frac{1}{R_0} && 0 && 0 && ... && 0 \\
    %     0 && \frac{1}{R_1} && 0 && ... && 0 \\
    %     && &&            ...\\
    %     0 && 0 && ... && \frac{1}{R_{N-2}} && 0 \\
    %     0 && 0 && ... && 0 && \frac{1}{R_{N-1}} \\
    % \end{bmatrix}
    \mathbf{A}=\text{diag}([\frac{1}{R_0}, \frac{1}{R_1}, ..., \frac{1}{R_{N-1}}])
\end{equation}

\begin{equation*}
  \small
    \mathbf{B}=
    \begin{bmatrix}
        \frac{-1}{3} && \frac{1}{3} && 0 && ... && 0 \\
        0 && \frac{-1}{3} && \frac{1}{3} && ... && 0 \\
        && &&            ...\\
        0 && 0 && ... && \frac{-1}{3} && \frac{1}{3} \\
        \frac{1}{3} && 0 && ... && 0 && \frac{-1}{3} \\
    \end{bmatrix}
\end{equation*}
% \end{align*}
% 
\begin{align*}
  \small
% \begin{equation*}
    \boldsymbol{\gamma}_1&=
    \begin{bmatrix}
        26.5 \\
        26.5 \\
        ...  \\
        26.5 \\
        26.5 \\
    \end{bmatrix}
% \end{equation*}
   &
    \boldsymbol{\gamma}_2&=
    \begin{bmatrix}
        a_{max}(v_0)\\
        a_{max}(v_1)\\
        ...\\
        a_{max}(v_{N-2})\\
        a_{max}(v_{N-1})\\
    \end{bmatrix}
    &
    \boldsymbol{\gamma}_3&=
    \begin{bmatrix}
        a_{min}(v_0)\\
        a_{min}(v_1)\\
        ...\\
        a_{min}(v_{N-2})\\
        a_{min}(v_{N-1})\\
    \end{bmatrix}
\vspace{-4mm}
\end{align*}

%We utilize the SciPy library's implementation of SQP to carry out the optimization.  
%This fits the standard definition of Sequential Quadratic Programming (SQP). 
We then train our neural network architecture to predict this raceline, $\mathcal{T^*}$, 2.25$[\second]$ into the future. 2.25$[\second]$ was selected as the duration of a typical braking zone for an F1 car. The input to this neural network is a sequence of images from the ego vehicle's point of view. Its output is a B\'ezier curve approximation of  $\mathcal{T^*}$, starting at a point on the raceline closest to the ego vehicle, as seen in Figure \ref{fig:pointestimates}. 
However, since this output is generated using supervised learning, it may not be correct as the network may encounter input image sequences which are outside of the training distribution. 
This network prediction serves to specify a prior distribution for our Bayesian filtering framework.
%We now describe that Bayesian framework.{\color{red}}{\color{red}}

\section{Differential Bayesian Filtering}
\label{sec:dbf}

% The neural network's estimate of $\mathcal{T^*}$ may not be correct, i.e. they represent an incorrect prior. 
% However, 
Since the \emph{differential} properties of $\mathcal{T^*}$ are well-defined, they can serve as the basis for inferring the optimal raceline given a prior.  Specifically, for the optimal raceline:

\begin{enumerate}
    \item $\mathbb{E}[\vert\dot{\mathcal{T^*}}\vert]$ (Eq.~\ref{eqn:optimal_raceline_definition}) is maximized.
    \item $\mathcal{T^*}$ respects the vehicle's dynamic constraints.
    \item The signed distance from $\mathcal{T^*}$ to the track bounds is strictly non-positive. ``Signed distance" means euclidean distance to the track boundary, but with a negative sign for points inside the track boundaries.
\end{enumerate}

The goal of Differential Bayesian Filtering is to infer a B\`ezier curve approximation of $\mathcal{T^*}$ by performing Bayesian estimation on $\mathcal{T^*}$'s differential properties, even if the global properties of $\mathcal{T^*}$ are not known.  In the context of our problem, this means that the resulting curve should have a higher average velocity, but still be within the track bounds and be physically achievable by the ego vehicle. 

Recall that we take the control points of B\`ezier curve to be the parameters, $\theta$, of our estimate of the optimal raceline: $\mathcal{T}_{\theta}$. I.e. $\theta=\{\mathbf{C}_0, \mathbf{C}_1, ..., \mathbf{C}_{k-1}\}$ as described in section \ref{sec:pbc}. We use a probabilistic B\`ezier curve, Equation~\ref{eqn:pbc_set} from Section~\ref{sec:pbc}, as a prior distribution, $p(\theta)$, and apply recursive Bayesian estimation with a likelihood function derived from $\mathcal{T^*}$'s differential properties to infer a more accurate estimate of true optimal behavior.  In general, for any curve that can be parameterized by $\theta$, Bayes' Theorem says:
\begin{equation}
    p(\theta \vert x) = \frac{l( \theta | x )p(\theta)}{\int l( \theta | x )p(\theta)d\theta}
    \label{eqn:bayes_generic}
\end{equation}
%{\color{red}}
The evidence, $x$, is sensor data or other state information that is available to the autonomous agent. For this work, we take this evidence to be the following:
\begin{enumerate}
\item A permissible offset distance from the track boundaries
\item The maximum allowed longitudinal acceleration.
\item The maximum allowed centripetal acceleration.
\end{enumerate}
$l( \theta | x )$ is a function of $\theta$ and $x$ that represents the likelihood that $\mathcal{T}_\theta$ (the trajectory defined by $\theta$) is the optimal raceline given the evidence, $x$.  The prior, $p(\theta)$, and likelihood function, $l( \theta | x )$, are the defining features of a Differential Bayesian Filtering model. 
Intuitively, this likelihood should be selected such that curves with \textit{desirable} differential properties have a higher likelihood and curves with \textit{undesirable} differential properties should have a lower likelihood, such that the posterior distribution, $p(\theta \vert x)$, represents a more accurate belief about $\mathcal{T^*}$. 
Other desirable racecar behaviors such as collision-free trajectories for multi-agent racing could be incorporated as another likelihood function in the future.%We now describe our prior and likelihood function.  } {\color{red}}
\subsection{Gaussian Prior on $\mathcal{T^*}$ }
\label{subsec:prior_spec}

For this work, we use a Probabilistic Bezier Curve as the basis for a prior distribution on the parameters of the optimal racing line.  We take the point-estimate output of our neural network model in \cite{deepracing_bezier} as the mean of our prior distribution and with identity covariance. I.e. our prior $p(\theta)$ is as follows:

\begin{equation}
    \small
    \{ \mathcal{N}(\mathbf{\hat{C}}_0, \begin{bmatrix} 1.0 & 0 \\ 0 & 1.0\end{bmatrix}) ,\ ...\ \mathcal{N}(\mathbf{\hat{C}}_k,\begin{bmatrix} 1.0 & 0 \\ 0 & 1.0\end{bmatrix}) \}
    \label{eqn:dbf_prior}
\end{equation}

Where $\{\mathbf{\hat{C}}_0\ ...\ \mathbf{\hat{C}}_k\}$ are the point-estimate control points of the curve predicted by our neural network. 
 %However, we make one key change. \hl{I think what you are saying here is that you do this change once to spin out the car, but the way the text is written implies that this scaling is a required step in your process. It may be prudent to remove this part and move it to the experiment section where you can refer to this subsection and equation number for context} We scale the time delta, $\Delta t$, that the neural network was trained to predict by a factor of $\frac{1}{1.15} \approx .870$. I.e. we scale up the velocities of the predicted curve by a factor of 15\%. Because centripetal acceleration is proportional to the square of velocity, this would increase the centripetal acceleration of the prior by a factor of ${1.15}^2 \approx 1.323$.  With no other changes to the desired trajectory, this would push the car beyond it's dynamic limits.  We show this empirically in section \ref{sec:results}.
%In order to bring the desired trajectory back with the vehicle's limits, 
Identity covariance is chosen for this initial work for simplicity, but other priors could be specified, e.g. where the covariance is determined based on the trajectories from the training data.
%{\color{red}}{\color{red}Identity covariance is chosen for this initial work for simplicity and to avoid the computationally expensive Cholesky decomposition required to sample from a non-diagonal covariance, but this model is extensible to other methods of specifying a Gaussian prior, e.g. one whose covariance is aligned to the tangent and lateral directions of the prior mean to generate more samples along one of those directions.}

%{\color{red}We also need to specify a likelihood function that discourages infeasible paths, such that the posterior is a better estimate of the optimal racing line.  We now describe that likelihood.}

\subsection{Likelihood Specification for $\mathcal{T^*}$ }
\label{subsec:likelihood_spec}
We want to specify our Bayesian model to make physically infeasible or out-of-bounds curves less likely and feasible curves that remain in-bounds more likely.  To achieve this goal, we utilize Bayes' Theorem where the likelihood function is a function of the \emph{differential properties} of the trajectory defined by $\theta$, rather than a function of only $\theta$:%. Expressed mathematically
\begin{equation}
    p(\theta \vert x) = \frac{l( \mathcal{T}_\theta, \dot{\mathcal{T}_\theta}, \ddot{\mathcal{T}_\theta} | x )p(\theta)}{\int l( \mathcal{T}_\theta, \dot{\mathcal{T}_\theta}, \ddot{\mathcal{T}_\theta} | x )p(\theta)d\theta}
    \label{eqn:bayes_dbf}
\end{equation}
%{\color{red} }

Note that under this formulation, there need not be a direct connection between the evidence, $x$, and the true parameters of $\mathcal{T}^*$. In our case, the likelihood is a function of the derivatives of $\mathcal{T}^*$.  To formally specify this likelihood, let $a_{c}$ and $a_{l}$ represent centripetal and longitudinal acceleration, respectively.  Let $d(\mathcal{T}_\theta)$ be the maximum signed distance from $\mathcal{T}_\theta$ to the track boundaries. In this context, ``signed distance" is Euclidean distance but with a negative sign for points on $\mathcal{T}_\theta$ that are inside the track bounds and a positive sign for points outside the bounds. We define acceleration  likelihoods (Eqs. \ref{eqn:l_acent}, \ref{eqn:l_along}) and a boundary likelihood (Eq. \ref{eqn:l_boundary}) as follows: %Figure \ref{fig:examplepaths} shows an illustration of this signed distance alongside examples of a curve with large centripetal acceleration.

% \vspace{-2mm}
\begin{flalign}
\noindent
\small
    l_{1}( \dot{\mathcal{T}_\theta}, \ddot{\mathcal{T}_\theta} ) = 
\begin{cases}
    e^{-\beta_{1}\Delta a_{c}(\dot{\mathcal{T}_\theta}, \ddot{\mathcal{T}_\theta})}, & \Delta a_{c}(\dot{\mathcal{T}_\theta}, \ddot{\mathcal{T}_\theta}) > 0\\
    1,                                            & \text{otherwise}
\end{cases}
\label{eqn:l_acent}
\end{flalign} 
\begin{flalign}
\noindent
\small
    l_{2}(\dot{\mathcal{T}_\theta}, \ddot{\mathcal{T}_\theta}) = 
\begin{cases}
    e^{-\beta_{2}\Delta a_{l}(\dot{\mathcal{T}_\theta}, \ddot{\mathcal{T}_\theta})}, & \Delta a_{l}(\dot{\mathcal{T}_\theta}, \ddot{\mathcal{T}_\theta}) > 0\\
    1,                                            & \text{otherwise}
\end{cases}
\label{eqn:l_along}
\end{flalign} 
% \vspace{-2mm}
\begin{flalign}
\noindent
\small
    l_{3}(\mathcal{T}_\theta \subset \mathcal{L}) = 
    \begin{cases}
        e^{-\beta_{3}(d(\mathcal{T}_\theta) - d_{min})},& d > d_{min}\\
        1,              & \text{otherwise}
    \end{cases}
\label{eqn:l_boundary}
\end{flalign}

In this context, $\Delta a_{c}(\dot{\mathcal{T}_\theta}, \ddot{\mathcal{T}_\theta})$ represent how much $\mathcal{T}_\theta$ violates the centripetal acceleration limits of the car; $\Delta a_{c}(\dot{\mathcal{T}_\theta}, \ddot{\mathcal{T}_\theta})>0$ implies the curve is too aggressive (too much centripetal acceleration) and $\Delta a_{c}(\dot{\mathcal{T}_\theta}, \ddot{\mathcal{T}_\theta})<=0$ implies the curve is within the limit. $\Delta a_{l}$ represents the same function, but for longitudinal acceleration (braking/throttling). These deltas refer to the same feasibility limits (linear and centripetal acceleration) that were used when determining the optimal raceline for training the neural network.  We refer to trajectories that exceed these acceleration limits as infeasible trajectories and to trajectories within the limits as feasible trajectories.  $\beta_{1}$, $\beta_{2}$, $\beta_{3}$, and $d_{min}$ are all tuneable hyperparameters.  These parameters represent how strongly each factor of the likelihood function should be weighted. 
% \vspace{-3mm}{\color{red}}
% \begin{flalign}
% \noindent
% \small
%     l_{a}(\dot{\mathcal{T}_\theta}, \ddot{\mathcal{T}_\theta}) = l_1(\dot{\mathcal{T}_\theta}, \ddot{\mathcal{T}_\theta})l_2(\dot{\mathcal{T}_\theta}, \ddot{\mathcal{T}_\theta})
% \label{eqn:l_a}
% \end{flalign}
% % \vspace{-6mm}
% \begin{flalign}
% \noindent
% \small
%     l_{b}(\mathcal{T}_\theta) = l_3(\mathcal{T}_\theta)
% \label{eqn:l_a}
% \end{flalign}
Our likelihood function for Bayesian inference is the product of these three factors:
\begin{equation}
    l( \mathcal{T}_\theta, \dot{\mathcal{T}_\theta}, \ddot{\mathcal{T}_\theta} | x ) = l_{1}( \dot{\mathcal{T}_\theta}, \ddot{\mathcal{T}_\theta} )l_{2}(\dot{\mathcal{T}_\theta}, \ddot{\mathcal{T}_\theta})l_{3}(\mathcal{T}_\theta \subset \mathcal{L})
    \label{eqn:overall_likelihood}
\end{equation}
The core idea is that, given a prior distribution that is at or beyond the limits of control, our likelihood model infers optimal behavior by encouraging feasible trajectories that are within bounds and discouraging infeasible or out-of-bounds trajectories. Our method is not limited to these definitions of feasibility or to these specific acceleration limits, they are just a simplifying choice.  More complex vehicle dynamics could also be included in our approach by specifying an appropriate likelihood function based on such a model, to account for slip angles, tire forces etc.
\subsection{Sampling-Based Differential Bayesian Filtering}
\label{subsec:sampling}

\begin{figure}%[h]
    \centering
    \includegraphics[width=.85\columnwidth]{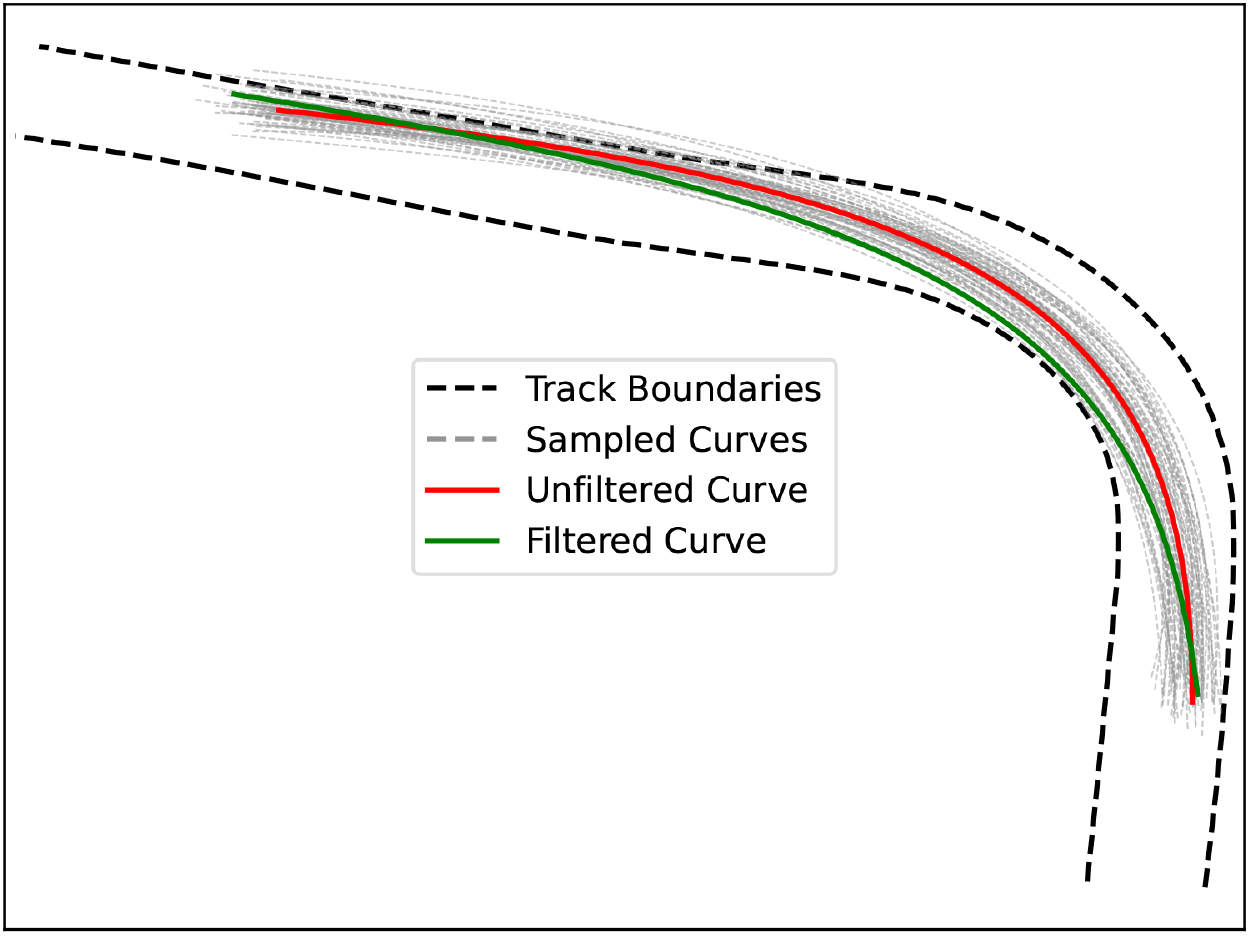}  
    \caption{One step of the Differential Bayesian Filtering algorithm. The prior (unfiltered) curve in red is infeasible, with too much centripetal acceleration. Our filtering approach produces a posterior (in green) that is closer to optimal with a softer arch that still maintains it's speed.}
    \label{fig:filteringexample}
\end{figure}

The closed-form solution to the posterior distribution under our formulation is not readily obvious. To remedy this, we now describe a sampling-based approach for generating samples from the posterior distribution without the need to solve for it analytically.  We employ Monte Carlo-based sampling from the posterior distribution over $\theta$ given the likelihood specified in subsection \ref{subsec:likelihood_spec} and the prior from subsection \ref{subsec:prior_spec}, $p(\theta)$. We draw a sample of N curves $p(\theta)$: $[\theta_1, \theta_2, \theta_3, ..., \theta_N]$ from $p(\theta)$.  These $N$ curves are then assigned a weight based on equations \ref{eqn:l_acent}, \ref{eqn:l_along}, and \ref{eqn:l_boundary}. This approach is extensible to more sophisticated sampling schemes such as Adaptive Monte Carlo, but a fixed number of samples is used in this work.
% {\color{red}}\hl{Where $Softmax$ is standard Softmax function common in machine learning:}
% $$Softmax(\beta_{v}*\Bar{v}(\theta)) = \frac{e^{\beta_{v}*\Bar{v}(\theta)}}{\sum_{i=1}^Ne^{\beta_{v}*\Bar{v}(\theta_i)}}$$
%Each $\theta_i$ is one of the N samples drawn from the prior distribution.  The high-level idea is that curves that exceed the vehicle's dynamic constraints or are too close to (or outside of) the track boundaries are discouraged (lower likelihood).
% \begin{figure}
%     \centering
%     % \includegraphics[width=0.7\columnwidth]{images/curvescartoon.png}
%     \includegraphics[width=0.675\columnwidth]{images/curvescartoon.pdf}
%     \caption{Some example trajectories. One is discouraged by our likelihood because of it's large centripetal acceleration, another because of an out-of-bounds violation, and a third (in green) is much closer to optimal.}
%     \label{fig:examplepaths}
% \end{figure}
We then employ a re-sampling step to infer a posterior distribution.  Each sample $\theta_i$ is assigned a weighting factor $\gamma_i$ according to:
\begin{equation}
    \alpha(\theta_i) = l_{1}( \dot{\mathcal{T}_{\theta_i}}, \ddot{\mathcal{T}_{\theta_i}} )l_{2}(\dot{\mathcal{T}_{\theta_i}}, \ddot{\mathcal{T}_{\theta_i}})l_{3}(\mathcal{T}_{\theta_i} \subset \mathcal{L})
    \label{eqn:alpha}
\end{equation}
\begin{equation}
\small
    \gamma(\theta_i) = \frac{\alpha(\theta_i)}{\sum_{i=1}^N\alpha(\theta_i)}
    \label{eqn:gamma}
\end{equation}
$l_1$, $l_2$, and $l_3$ refer to equations \ref{eqn:l_acent}, \ref{eqn:l_along}, and \ref{eqn:l_boundary}, respectively.  The mean of the posterior distribution is then taken as $\sum_{i=1}^N\gamma_i*\theta_i$. %The covariance of this Gaussian posterior is taken as the sample covariance matrix of another N $I.I.D.$ samples from a discrete random variable with probability mass function $\{ (\theta_i, \gamma_i)\text{, } 1\leq i \leq N \}$.  
This process repeats for a desired number of iterations. Algorithm 1 describes the high-level loop of our approach. Figure~\ref{fig:filteringexample} depicts our approach graphically. After sampling the grey curves and assigning their weights, $\gamma_i$, the posterior mean (in green) is inside the track bounds and is dynamically feasible while maintaining a high speed.%{\color{red}}

\noindent \textbf{Novelty of DBF compared to Behavioral Cloning approaches:} The novelty of the DBF method is the following: (1) Instead of tuning a neural network for predicting a single parameterized trajectory as one would in a behavioral cloning setting, we view the problem as one of learning probability distributions over the parameters of the optimal trajectory; (2) At runtime, we sample from the learned distributions to create candidate trajectories; (3) Using our likelihood specification, we emphasize samples with desirable properties and discourage samples with undesirable properties; (4) Finally, we use a weighting factor based on this likelihood to generate a posterior distribution that exhibits more optimal behavior.
This approach differs from the typical view of a machine-learned trajectory synthesis model by incorporating knowledge of the vehicle's dynamics to improve the point-estimate produced by a deep model.

\vspace{-5mm}

\section{Experiments \& Results}
\label{sec:results}
We now present an empirical evaluation of Differential Bayesian Filtering in a high-speed, Formula One racing environment. %{\color{red}}
We use the neural network architecture from our previous work in~\cite{deepracing_bezier} to provide a point estimate of a B\`ezier Curve approximation to the optimal racing line, $\mathcal{T}^*$, given a sequence of images taken from the driver's point-of-view, as shown in Figure~\ref{fig:pointestimates}. This point estimate serves as the mean of our prior distribution, but with one key change. We scale the time delta, $\Delta t$, that the neural network was trained to predict by a factor of $\frac{1}{1.15} \approx .870$. I.e. we scale up the velocities of the prior by a factor of 15\%. Because centripetal acceleration is proportional to the square of velocity, this would increase the centripetal acceleration of the prior by a factor of ${1.15}^2 \approx 1.323$.  With no other changes to the trajectory, this would push the car beyond its feasibility limits (we show this empirically in subsection \ref{subsec:dynamics}).%{\color{red}}  
Even though this scaled trajectory is infeasible, we use this as a prior - meaning that the mean of the distribution is dynamically infeasible and will cause the racecar to spin out.  This unstable prior is chosen in order to encourage the Bayesian sampling to produce candidate trajectories which are a mix of both stable and unstable raceline estimates. Consequentially, this increases the probability of finding a trajectory that is close to the vehicle's limits, but does not exceed them.%{\color{red}}{\color{red}}
Differential Bayesian Filtering improves this initial unstable prior into a more optimal B\`ezier curve that is both dynamically feasible and faster than the neural networks point estimate.
The resulting optimal B\'ezier curve is then passed to a Pure Pursuit~\cite{pure_pursuit} controller to generate steering and throttle commands for the autonomous racing agent. This Pure Pursuit controller uses an adaptive lookahead based on a control law of $l_d=0.4v$ with $l_d$ and $v$ representing lookahead distance and the car's current speed, respectively. We evaluate the unfiltered approach, where just the point estimate off the neural network (with no velocity scaling) is passed straight to Pure Pursuit, as well as a filtered model with Differential Bayesian Filtering on the unstable prior to produce a posterior that is within the car's limits. 
%Additionally, we show that the mean of our prior distribution (with the scaled up velocity but no filtering) is dynamically infeasible and results in unacceptable centripetal acceleration causing the racecar to spin out.{\color{red}}

In addition, we also compare our approach to two other models that uses a behavioral cloning approach to mimic an expert human driver's behavior instead of following the optimal racing line. One of these models was trained to predict a B\`ezier Curve that mimics the trajectory taken by the expert human driver. The other is closer to Waymo's ChauffeurNet\cite{ChauffeurNet} and predicts a series of waypoints followed by the expert example. We also include performance results when the Pure Pursuit controller is given ground-truth knowledge of the SQP-generated raceline the network was trained on.

Each model is run for 5 laps in our DeepRacing framework, using Codemaster's F1 2019 racing video game and the DeepRacing closed-loop autonomous racing infrastructure~\cite{deepracing-date}.  On each run, we measure the following metrics:
\begin{enumerate}
    \item Overall Lap Time
    \item Average Speed
    \item How many times the autonomous agent went outside the track bounds, what we call a ``Boundary Failure" - Such a failure is defined as when $\geq3$ of the vehicle's tires go out of bounds.
    % \item The mean time between Boundary Failures
\end{enumerate}
%This specifies our prior as in section \ref{subsec:prior_spec}. We then test each model by running 5 laps on Albert Park Circuit in the F1 game and measuring all of the specified metrics.
In these experiments, order 7 curves were chosen (k=7) based on heuristics. Table \ref{table:hyperparameters} shows the values we use for all of the hyperparameters for our approach defined in section \ref{sec:dbf}.
Table \ref{table:empirical_results} summarizes the performances of DBF in the DeepRacing simulator (means across 5 laps). DBF has the fastest overall lap time and the fastest average speed.  This implies our method is taking a more efficient racing line and is doing so more aggressively without exceeding dynamic limits of the car.  It is also worth noting that our method has no boundary failures.  Our approach also improves lap time by $\sim4.458[\second]$ on average from the unfiltered neural network's raceline predictions. In the world of high-speed racing, where winners/losers can be decided by millisecond time differences, this is a very significant improvement. We also achieve lap times superior to the human driving examples used to train the neural network by $\sim2.4[\second]$.  I.e. our technique pushes the vehicle faster than even the fastest example in the neural networks training distribution.  All of our experiments were conducted on a PC with 12 CPU cores and an NVIDIA GTX1080Ti GPU.  Each iteration of the filtering loop (250 samples) required an average of $0.034[\second]$ of computation time, implying a rate of $\sim29[\hertz]$ for each iteration of DBF.  Our overall algorithm, including a call to AdmiralNet and the pure pursuit controller, runs at $10[\hertz]$.

\begin{table}%[]
\caption{Hyperparameters used for our experiments}
\centering
\resizebox{.75\columnwidth}{!}{%
\begin{tabular}{|c|c|}
\hline
\textbf{Hyperparameter}                                                                   & \textbf{Value} \\ \hline
\begin{tabular}[c]{@{}c@{}}$\beta_1$ (Eqn \ref{eqn:l_acent})\end{tabular}  & 1.75          \\ \hline
\begin{tabular}[c]{@{}c@{}}$\beta_2$ (Eqn \ref{eqn:l_along})\end{tabular} & 2.5   \\ \hline
\begin{tabular}[c]{@{}c@{}}$\beta_3$ (Eqn \ref{eqn:l_boundary})\end{tabular}  & 3.5   \\ \hline
\begin{tabular}[c]{@{}c@{}}$d_{min}$ (Eqn \ref{eqn:l_boundary})\end{tabular} & -0.875 meters \\ \hline
\begin{tabular}[c]{@{}c@{}}$N$ (\# of samples in \ref{subsec:sampling})\end{tabular} & 250 \\ \hline
\end{tabular}%
}
\label{table:hyperparameters}
% \vspace{-3mm}
\end{table}
%Figure \ref{fig:speeds} also shows a Gaussian fit of the average speeds of both the filtered and unfiltered models.  Differential Bayesian Filtering results in faster racing behavior with less variance in the overall speed. Also, because the optimal racing line is often very close to the track bounds, following it increases the risk of boundary violations.  Differential Bayesian Filtering reduces this risk significantly, bringing it closer to the average number of boundary failures seen with a more conservative behavioral cloning approach. 
% \begin{figure}[!htb]
%     \centering
%     \includegraphics[width=0.85\columnwidth]{images/resultsplots/resultstable.pdf}
%     % \includegraphics[width=\columnwidth]{images/resultsplots/resultstable.png}
%     \caption{Empirical results from the Formula One simulation.  We also include results from our previous work for comparison.  Note that our approach (in bold and underlined) produces the best lap times and average speed.}
%     \label{fig:empirical_results}
% \end{figure}
% \include{results_fig.tex}
% \vspace{-3mm}
\begin{table}[!htb]
\caption{Results from experiments in the F1 simulator} 
% \vspace{-2mm}
\centering
\resizebox{\columnwidth}{!}{%
\begin{tabular}{|c|ccc|}
\hline
\textbf{\begin{tabular}[c]{@{}c@{}}Model\\ Configuration\end{tabular}} & \textbf{\begin{tabular}[c]{@{}c@{}}Lap \\ Time\\ $[\second]$\end{tabular}} & \textbf{\begin{tabular}[c]{@{}c@{}}Overall \\ Speed\\ $[\meter\per\second]$\end{tabular}} & \textbf{\begin{tabular}[c]{@{}c@{}}Number of\\ Boundary\\  Failures\end{tabular}} \\ \hline
\begin{tabular}[c]{@{}c@{}}Waypoint Prediction\\ (Behavioral Cloning)\end{tabular} & 106.683 & 49.245 & 5.6 \\ \hline
\begin{tabular}[c]{@{}c@{}}Bézier Curve Prediction\\  (Behavioral Cloning)\end{tabular} & 101.72 & 52.193 & 1.8 \\ \hline
\begin{tabular}[c]{@{}c@{}}Unfiltered Raceline \\ Prediction\end{tabular} & 91.219 & 58.109 & 0 \\ \hline
\begin{tabular}[c]{@{}c@{}}Fastest Human Lap\\ (Training Data)\end{tabular} & 89.177 & 59.439 & 1 \\ \hline
\begin{tabular}[c]{@{}c@{}}Ground-Truth Raceline\end{tabular} & 88.046 & 60.186 & 0 \\ \hline
\textbf{\begin{tabular}[c]{@{}c@{}}Raceline Prediction w/\\ Differential Bayesian Filtering (Ours)\end{tabular}} & {\ul \textbf{86.761}} & {\ul \textbf{61.092}} & {\ul \textbf{0}} \\ \hline
\begin{tabular}[c]{@{}c@{}}Nicholas Latifi\\ (in real-life 2019 Australian Grand Prix)\end{tabular} & 86.067 & Unknown & 0 \\ \hline
\begin{tabular}[c]{@{}c@{}}Lewis Hamilton\\ (in real-life 2019 Australian Grand Prix)\end{tabular} & 80.486 & Unknown & 0 \\ \hline

\end{tabular}
}
\label{table:empirical_results}
\vspace{1mm}
\small \newline Differential Bayesian Filtering outperforms the other methods and the best human lap in the training data by a significant margin.  Our method is also very close to competing with real F1 drivers.
% \vspace{-4mm}
\end{table}
% \vspace{-4mm}

\begin{figure}%[htb]
    \centering
    \includegraphics[width=0.95\columnwidth]{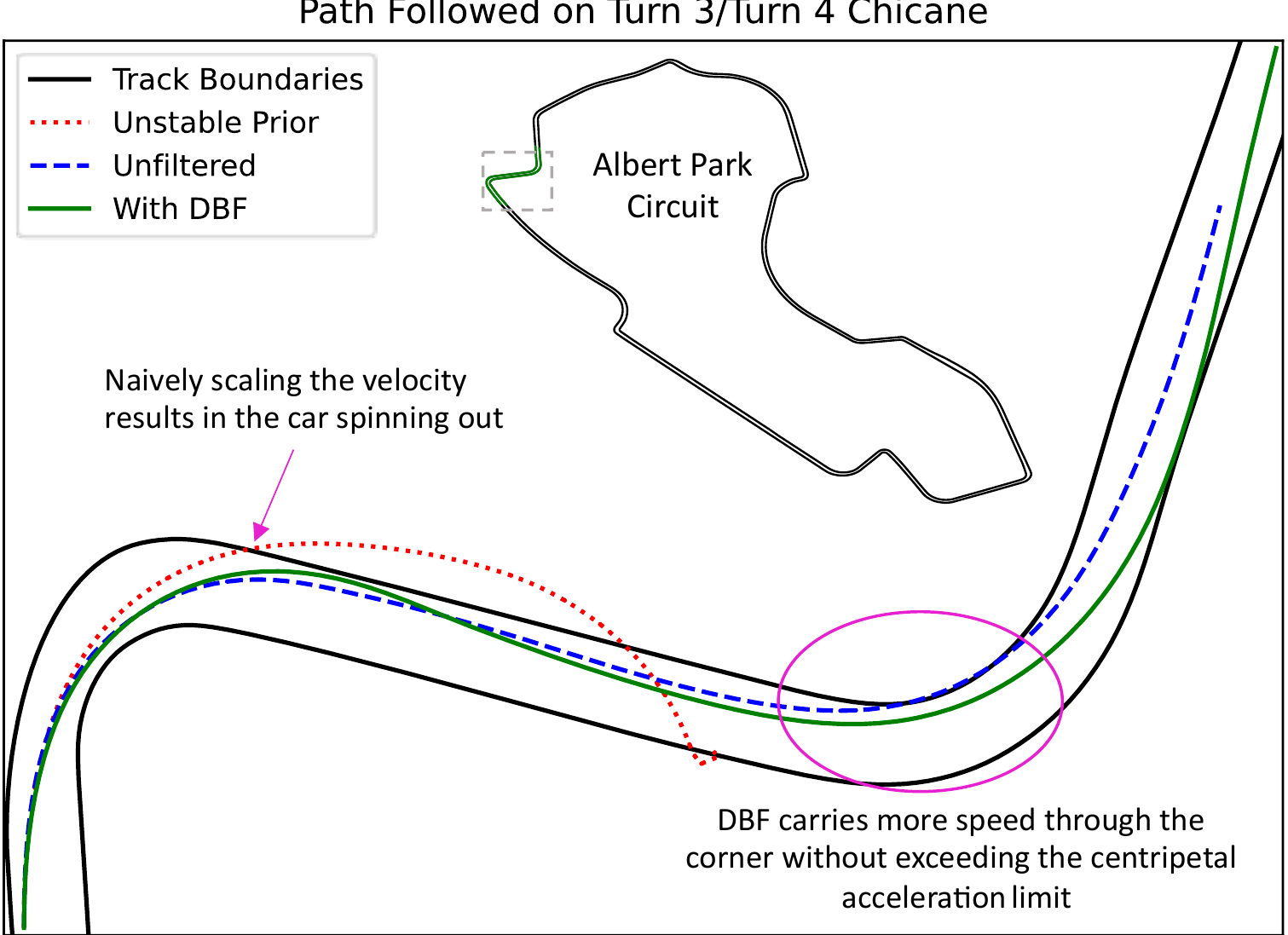}
    \caption{Turn3/Turn4 Chicane of Albert Park Circuit. Note that our methods results in a faster path (green) that carries more speed through the turn. The unstable prior (red) is not physically achievable and results in a crash.}
    \label{fig:path_comparison}
    % \vspace{-4mm}
\end{figure}
\subsection{Dynamics Analysis}
\label{subsec:dynamics}
\begin{figure}%[!htb]
    \centering
    \includegraphics[width=1.0\columnwidth]{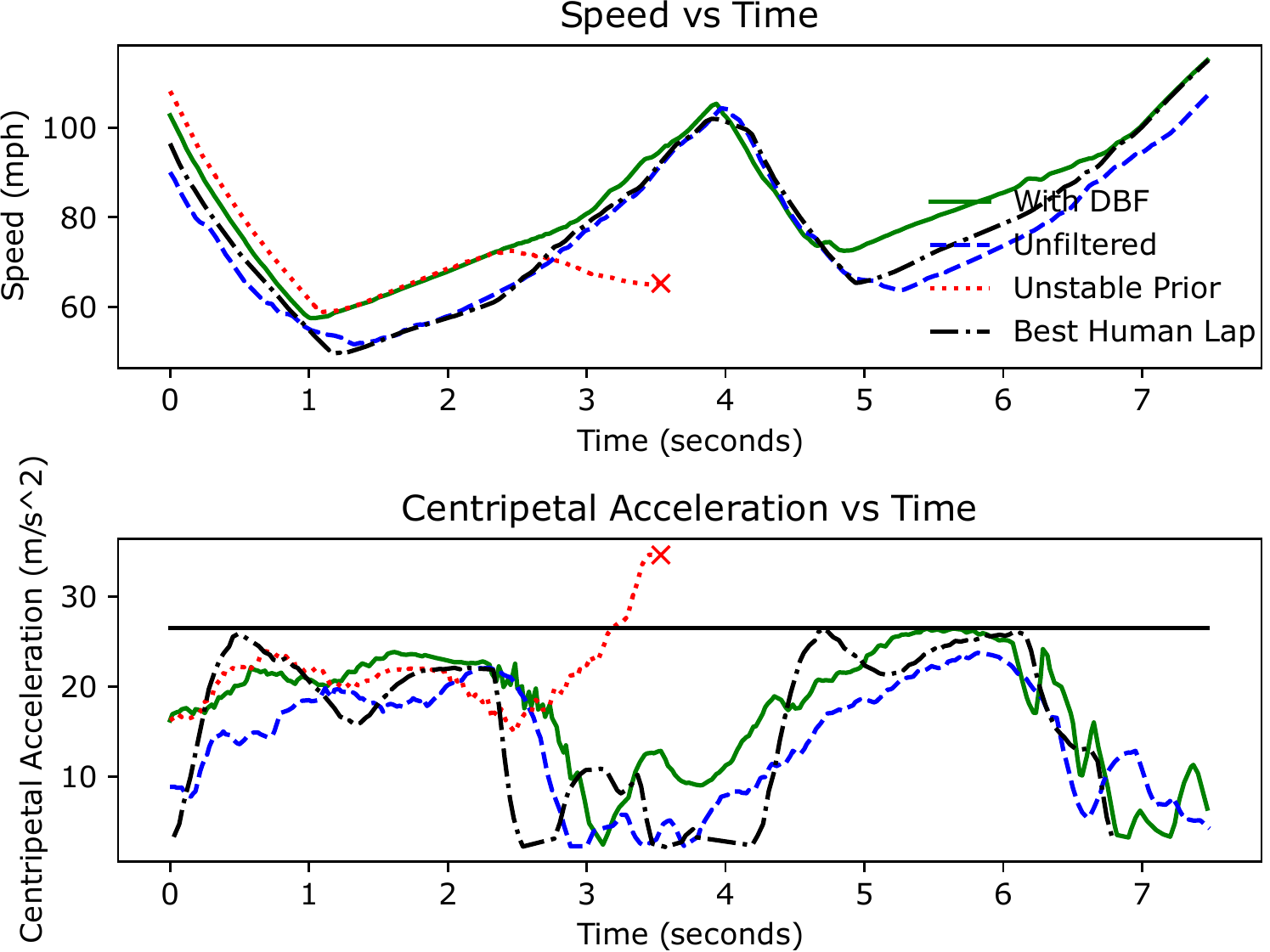}
    \caption{Our artificially accelerated prior is dynamically infeasible and causes the car to spin out on the Turn3-Turn4 chicane. Differential Bayesian Filtering brings that prior back within the car's physical limits. Our specified dynamic limit of $26.5[\meter\per\second\squared]$ is marked on the lower plot.  Our approach pushes the car right up to its dynamic limit, but does not exceed it. For the unstable prior, both plots end at the point the car spins out.}
    \label{fig:speed_centripetalaccel_comparison}
    % \vspace{-1mm}{\color{red}}{\color{red}}
\end{figure}

We also present evidence that our method is achieving our stated goal, pushing the vehicle its limits.  The Turn 3/Turn 4 chicane of Albert Park Circuit is a particularly challenging turn.  It requires braking from almost max speed and navigate an S-shaped chicane.  Fig \ref{fig:path_comparison} shows this chicane and how DBF (green curve) maintains speed through the turn by utilizing the available track width.  Fig \ref{fig:speed_centripetalaccel_comparison} shows the centripetal acceleration the car experiences in this chicane while controlled by each of 3 trajectory planners:
\begin{enumerate}
    \item The unfiltered neural network predictions 
    \item The artificially accelerated (unstable) prior distribution
    \item Our Differential Bayesian Filtering method
    \item The best human example lap from our training set
\end{enumerate}

The horizontal line shows the dynamic limit we assigned to the car. Note that when following the neural network's unfiltered predictions (dashed blue in Figure \ref{fig:path_comparison}), the car drives too passively.  There is significant room for the vehicle to drive faster without exceeding the vehicle's dynamic limits.  Also note that when pushed faster artificially (as with our selected prior, dotted red in Figure \ref{fig:path_comparison}), the car sees far too much centripetal acceleration (see Figure \ref{fig:speed_centripetalaccel_comparison}, even human drivers don't push the car that hard).  The vehicle becomes unstable and spins out halfway through the chicane under these conditions.
Our method results in the best of both worlds. The vehicle operates right at its dynamic limits, but does not exceed them.  The vehicle carries more speed through the turns, but with an acceptable level of centripetal acceleration.  This manifests as overall faster lap times.

In sum, we show that Differential Bayesian Filtering can produce a higher-performance racing agent that drives more aggressively (better lap time \& speed) and more safely (not going out-of-bounds) than only following point estimates of the optimal racing line produced by a convolutional neural network.  This technique is also generalizable to more application-specific likelihood models and prior distributions.  
% \subsection{}

\section{Conclusion}
\label{sec:conclusion}
This paper presents Differential Bayesian Filtering, a novel Bayesian framework for trajectory synthesis in high-speed autonomous racing situations.  Our method can be used to infer optimal trajectories from known differential properties of the optimal racing line by framing the problem as recursive Bayesian estimation on the control points of a B\`ezier curve.  We also present a sampling-based Monte Carlo method for DBF on B\`ezier curves. We evaluate the performance of DBF using our Formula 1 simulator, and show that it results in the overall best performance, compared to other approaches and the expert driving examples, on a variety of racing metrics. Future work includes extending this approach to a multi-agent racing setup as well changing the input space of the network from pixels of images to a canonical view of autonomous driving (e.g. the current state of the ego and any other agents) as the input space.

%Although a Gaussian prior with fixed variance was used in these experiments, our methodology is independent of this choice.  Future work would include a more principled selection of prior distribution based on domain knowledge and/or context-specific assumptions of the problem space as well as extending the approach to multi-agent settings.  

%\clearpage
%\newpage
% \bibliographystyle{unsrt}
% \bibliography{references,cps}

\end{document}